\documentclass[sigconf]{acmart}

\usepackage{algorithm}
\usepackage{algpseudocode}
\usepackage{tikz}
\usepackage{xcolor, colortbl}
\usepackage{balance}

\AtBeginDocument{%
  \providecommand\BibTeX{{%
    \normalfont B\kern-0.5em{\scshape i\kern-0.25em b}\kern-0.8em\TeX}}}

\copyrightyear{2022} 
\acmYear{2022} 
\setcopyright{rightsretained} 

\acmConference[ICMR '22]{Proceedings of the 2022 International Conference on Multimedia Retrieval}{June 27--30, 2022}{Newark, NJ, USA.}
\acmBooktitle{Proceedings of the 2022 Int'l Conference on Multimedia Retrieval (ICMR '22), June 27--30, 2022, Newark, NJ, USA}
\acmDOI{10.1145/3512527.3531366}
\acmISBN{978-1-4503-9238-9/22/06}

\usepackage{etoolbox}
\makeatletter
\patchcmd{\maketitle}{\@copyrightpermission}{
  \begin{minipage}{0.3\columnwidth}
     \href{http://creativecommons.org/licenses/by/4.0/}{\includegraphics[width=0.90\textwidth]{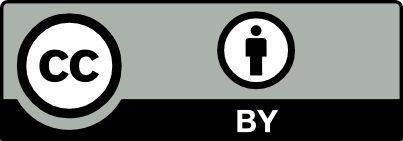}}
  \end{minipage}\hfill
  \begin{minipage}{0.7\columnwidth}
     \href{http://creativecommons.org/licenses/by/4.0/}{This work is licensed under a Creative Commons Attribution International 4.0 License.}
  \end{minipage}
  
  \vspace{5pt}
}{}{}

\makeatother

\begin{document}

\fancyhead{}
\title{Local Slot Attention for Vision-and-Language Navigation}

\author{Yifeng Zhuang}
\email{19210240018@fudan.edu.cn}
\affiliation{
    \institution{Fudan University}
    \country{China}
}

\author{Qiang Sun}
\email{18110860051@fudan.edu.cn}
\affiliation{
    \institution{Fudan University}
    \country{China}
}

\author{Yanwei Fu}
\email{yanweifu@fudan.edu.cn}
\affiliation{
    \institution{The school of Data Science, Fudan University}
    \country{China}
}

\author{Lifeng Chen}
\email{chenlf@fudan.edu.cn}
\affiliation{
    \institution{Fudan University}
    \country{China}
}
\authornote{Corresponding authors}

\author{Xiangyang Xue}
\email{xyxue@fudan.edu.cn}
\affiliation{
    \institution{Fudan University}
    \country{China}
}


\renewcommand{\shortauthors}{Zhuang, et al.}

\begin{abstract}
Vision-and-language navigation (VLN), a frontier study aiming to pave the way for general-purpose robots, has been a hot topic in the computer vision and natural language processing community. The VLN task requires an agent to navigate to a goal location following natural language instructions in unfamiliar environments. 

Recently, transformer-based models have gained significant improvements on the VLN task. Since the attention mechanism in the transformer architecture can better integrate inter- and intra-modal information of vision and language.

However, there exist two problems in current transformer-based models. 

1) The models process each view independently without taking the integrity of the objects into account. 

2) During the self-attention operation in the visual modality, the views that are spatially distant can be inter-weaved with each other without explicit restriction. This kind of mixing may introduce extra noise instead of useful information. 

To address these issues, we propose 
1) A slot-attention based module to incorporate information from segmentation of the same object.
2) A local attention mask mechanism to limit the visual attention span. 
The proposed modules can be easily plugged into any VLN architecture and we use the Recurrent VLN-Bert as our base model. Experiments on the R2R dataset show that our model has achieved the state-of-the-art results.
\end{abstract}




\begin{teaserfigure}
  \centering
  \includegraphics[width=0.86\textwidth]{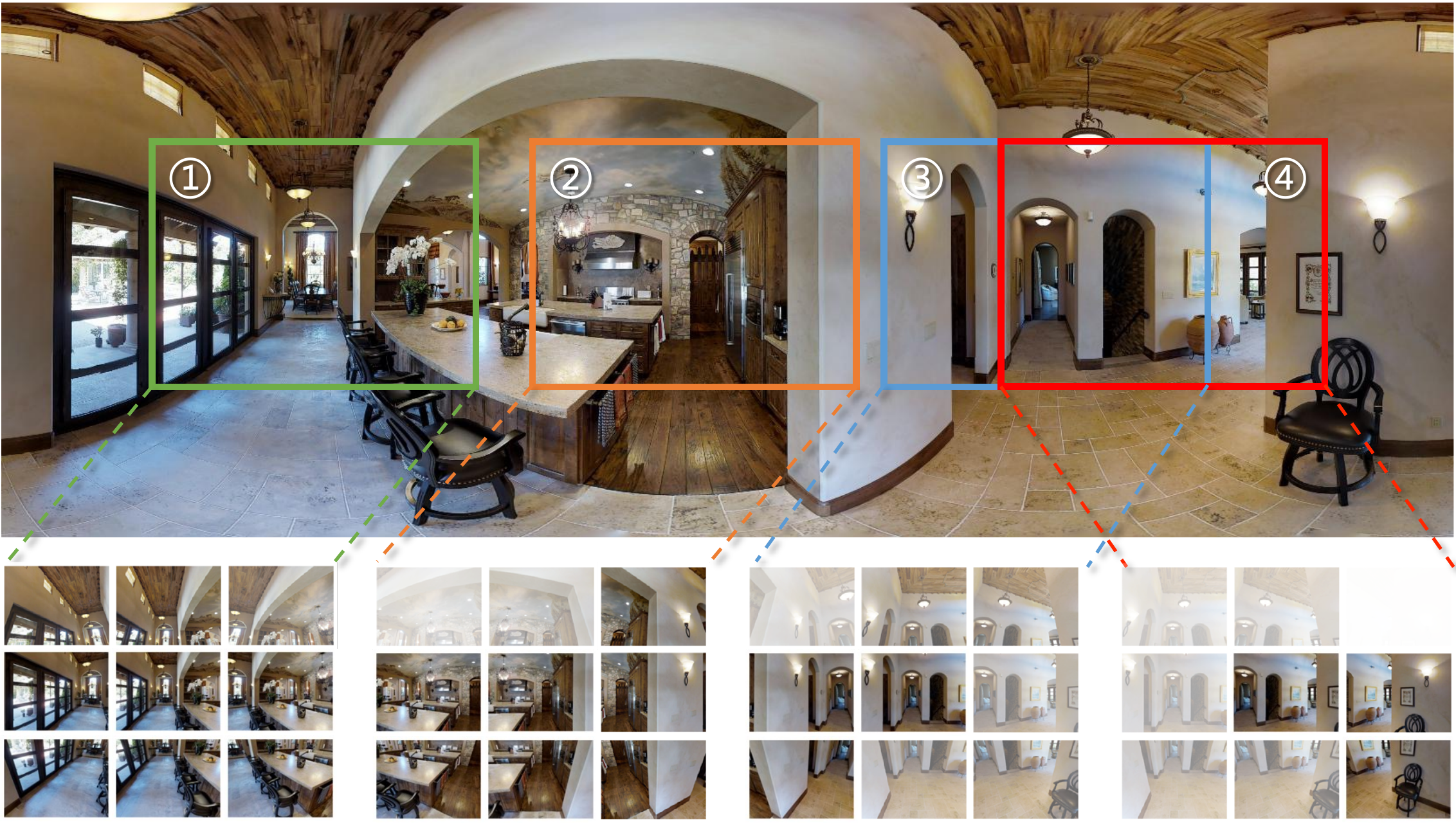}
  \caption{Illustration of our local slot attention module. In the top line is the panoramic view of the current viewpoint. The colored boxes represent candidate views. In the bottom line are 3x3 grids with candidate views in the center, and the candidate views can only attend to those views around them inside the grid. Higher opacity indicates higher attention weight. }
  \Description{Candidate views can attend their surrounding views with different weights, and the context views help the candidate views to rebuild the information of partially missing objects.}
  \label{fig:teaser}
\end{teaserfigure}

\maketitle


\section{Introduction}

\begin{figure*}[!htbp]
  \centering 
  \includegraphics[width=0.9\linewidth]{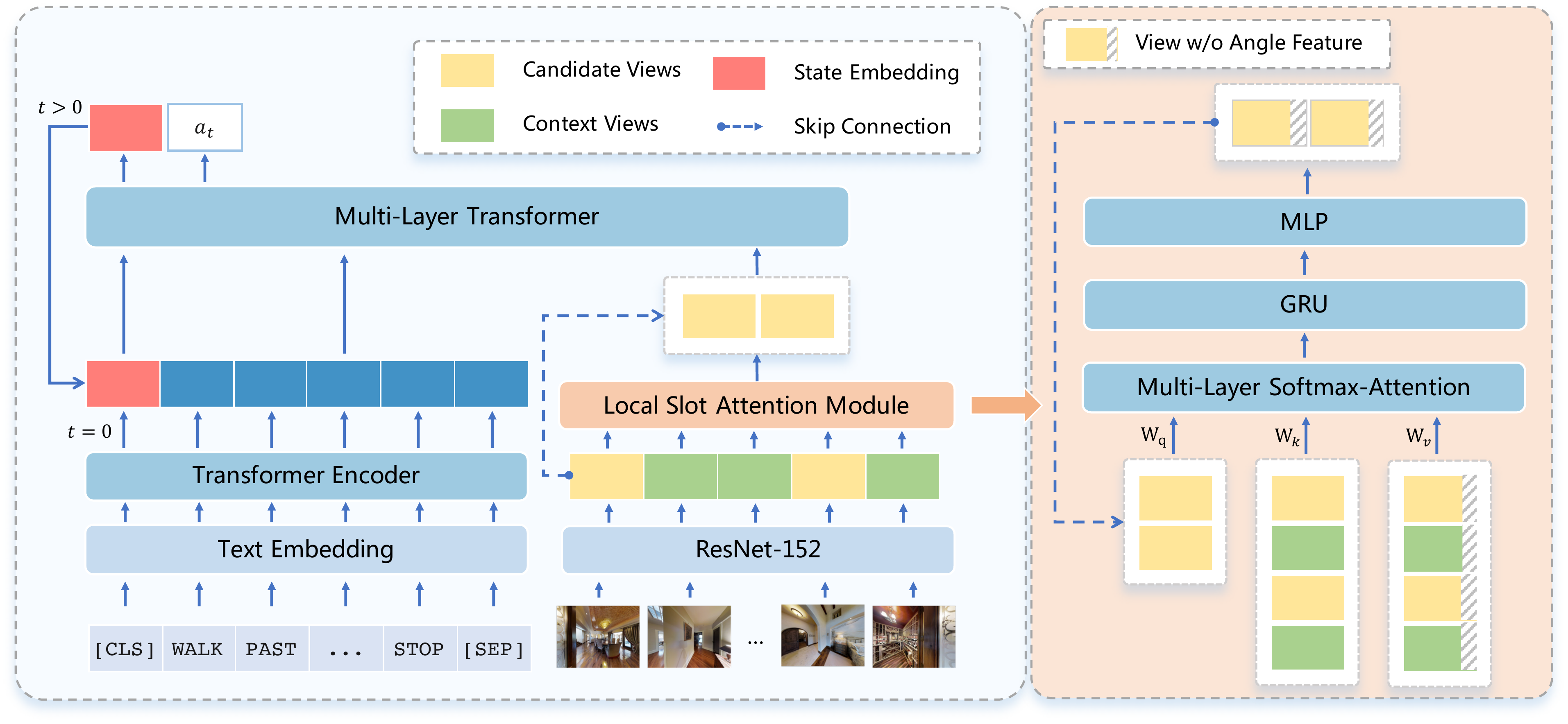}
  \caption{On the left is the overview of our model. On the right is the local slot attention module we propose. We only illustrate one iteration in the local slot attention module for a clear demonstration.}
  \Description{An overview of our proposed model.}
  \label{fig:model}
\end{figure*}
It is a long-term goal to develop a general-purpose servant robot that can understand natural language instructions. Vision-and-language navigation (VLN), which demands an agent to follow instructions and navigate in previously unseen indoor environments, has been a good testbed towards this goal. During one round of navigation, an agent is provided with detailed instructions such as "Walk out of the washroom past the double closet doors and walk into the next room.", and the agent must follow these instructions and make proper action decisions. 
 
Early works \cite{anderson2018vision, tan2019learning} are built on RNN-based architectures that model the VLN task as a sequence-to-sequence problem. In the speaker-follower~\cite{fried2018speaker} model, a general setting of panoramic action space is proposed which represents each viewpoint as a panorama composed of 36 views of different headings and elevations; and this setting is adopted by most of the current works. To reinforce the alignment between vision and language, the PREVALENT~\cite{hao2020towards} was proposed as the first transformer-based model that encodes inputs of different modalities as a joint representation by utilizing the vision-language pre-training, and it boosts the performance on the R2R dataset by a large margin. The Recurrent VLN-Bert~\cite{Hong2021vlnbert} further introduces a recurrent function to model the navigation history and achieves the state-of-the-art while improving the training efficiency. 

However, these works~\cite{hao2020towards, Hong2021vlnbert} apply self-attention on visual inputs in a counter-intuitive way. As formulated in PREVALENT~\cite{hao2020towards}, all 36 views of one viewpoint will attend to each other at each time step to integrate information from the whole scene. This is not only costly in computation but also against common sense, as the regions with large spatial distances are usually unrelated. Such information from remote views can become noises that harm the navigation. In Recurrent VLN-Bert~\cite{Hong2021vlnbert}, only candidate views will be used for attentions. This reduces the memory usage at the cost of restricting the agent's visual perceptive field to isolated candidate views. However, visual clues can be cropped or obscured in candidate views and fail to correspond to language instructions.

To alleviate these problems, we propose a local slot attention module \textbf{LSA}, which includes:

1) A modified slot-attention-based module that enhances the features of candidate views with information from panoramic views (represented as 36 views). As illustrated in Fig.\ref{fig:teaser}, we treat each candidate view (bounded with colored boxes in the top line) as a slot that aims to capture information of a specific navigable region. All 36 views are treated as contextual views that will fill into these slots in a soft manner to incorporate global information into the local candidate views. 

2) A local attention mask mechanism that explicitly limits the attention span to be within a local region, preventing candidate views from incorporating noises from distant views. As shown in the bottom line in Fig. \ref{fig:teaser}, we apply a 3x3 mask that restricts the attention span of each slot to be within its closest neighbors.

To summarize our contributions: We propose a novel attention-based module to allocate information from panoramic views to candidate views in a soft manner, which helps to relives the loss of visual information in candidate views. 
Moreover, we introduce a local attention mask that helps the agent to focus on the most important local region around each candidate view. Our \textbf{LSA} module is easily plugged into any VLN model to achieve a significant performance boost while retaining a reasonable computational cost at the same time.

We demonstrate that our proposed \textbf{LSA} module helps to restore the integrity of the local information around candidate views. We validate this by applying \textbf{LSA}  into the Recurrent VLN-Bert\cite{Hong2021vlnbert} model. Results show that we can outperform the state-of-the-art on the R2R dataset within very few training iterations. The overview of our model is illustrated in Fig.\ref{fig:model}. Our code and pre-trained model is released on GitHub\footnote{https://github.com/PatZhuang/LSA}.


\section{Related Work}

\subsection{Vision-and-language Navigation}

VLN is a cross-modal task which combines the vision, language and action decision\cite{anderson2018vision, chen2019touchdown, krantz2020beyond, ku2020room, nguyen2019vision-based, nguyen2019help}. In this task, an agent is asked to navigate in visual environments following language instructions.
Recent works widely adopted the panoramic action space setting which proposed in the speaker-follower\cite{fried2018speaker}. Specifically, the visual environments are represented as discrete connectivity graphs, and the agent needs to choose one linked node to move to, or choose to stop at each time step, where each node represents a viewpoint in the scene. Once the agent moves to a new viewpoint, it perceives a panoramic visual input of this viewpoint which comprised of 36 views of different headings and elevations.
There are two lines of works in the VLN task, one is to use data augmentation to generate more training data, and the other line is to develop a more suitable model architecture for the VLN task. Our work belongs to the latter.

Data augmentation plays an important role in the VLN task, as the training data is very limited due to expensive annotation work. To alleviate this shortage, speaker-follower\cite{fried2018speaker} used a self-supervised training paradigm to generate new instruction-trajectory pairs from randomly sampled paths. Env-Drop\cite{tan2019learning} utilized an environmental consistent dropout scheme to augment visual inputs. 

For the model architecture, the RNN-based model~\cite{anderson2018vision} is first used as a strong baseline in the VLN task that encodes language instructions and decodes actions.
Recently, inspired by the success of BERT~\cite{devlin2019bert}, transformer-based models~\cite{federico2021perceive,hao2020towards, Hong2021vlnbert, qi2021road} have been proposed to address the VLN task. Results show that the transformer-based models can certainly out-perform the traditional RNN-based models, thanks to the attention mechanism that better aligns the information between vision and language. One remaining problem is that transformer-based models require high computational cost. To alleviate this, some works\cite{Hong2021vlnbert} limit the visual inputs to candidate views only, resulting in the loss of rich contextual information in background views. In our work, we achieve a balance between memory consumption and information integration via our proposed slot attention module with local attention mask mechanism.

\subsection{Transformer}

Transformer~\cite{vaswani2017attention} has been widely applied in the vision and language tasks, such as image caption~\cite{li2019entangled}, visual question answering~\cite{lu2019vilbert,tan2019lxmert} and vision-and-language navigation. With the well-designed pre-training tasks on large-scale datasets, the transformer can better align the vision and language modality and generalize well in downstream tasks. Specifically, PREVALENT\cite{hao2020towards} is the first transformer-based model on the VLN task that utilizes masked language modeling and action prediction as pre-training tasks. The Recurrent VLN-Bert~\cite{Hong2021vlnbert} replaces the LSTMCell in PREVALENT\cite{hao2020towards} that encodes the navigation state with a recurrent function to get rid of explicit usage of a recurrent module for efficiency. Our work is based on the original architecture of Recurrent VLN-Bert\cite{Hong2021vlnbert}.

\subsection{Slot Attention}

Slot attention is proposed in \cite{FrancescoLocatello2020ObjectCentricLW} that aims to learn object-centric representations. In slot attention, a set of $K$ vectors called slots are randomly initialized, these $K$ slots will query the $N$ input feature maps in multiple rounds of attention to refine themselves. During each iteration of attention, a soft-max function is applied over all slots, forcing each slot to compete for input feature maps. Finally, these slots will learn to divide the input into several parts and bind to different objects (or the background) in the input in an unsupervised manner. Moreover, such slots can be irrelevant to any specific object but work like object files that capture different types of objects from the input image. So the slot attention mechanism is capable of dealing with different numbers of slots, and different composition of (event unseen) objects.

Similarly, in the VLN task, we have different numbers of candidate views at each time step, and different visual characteristics among buildings. So we inherit the architecture of the slot attention module to build a modified version that uses candidate views as slots to query the whole panoramic view at each viewpoint.



\section{The VLN Task}

The VLN task is a sequential decision making task where the agent at time $t$ is required to make action decision $a_t$ based on a given language instruction $\mathbf{L}$ and the historical visual observations $\{{\mathbf{V}_i}\}_{i=0}^{t-1}$.
\begin{equation}
    a_t = \mathrm{argmax}_{a \in A_t} p(a|\{{\mathbf{V}_i}\}_{i=0}^{t-1}, \mathbf{L})
    \label{action_decision}
\end{equation}
where $A_t$ is a set of valid actions. After the agent executes the action $a_t$, it leaves the current viewpoint to another corresponding to $a_t$, and gains a reward $r_t$. The internal state $s_{t}$ of this round of navigation will be updated consequently. A decision process of $T$ steps can be denoted as follows.
\begin{equation}
    {s_0, a_1, r_1, s_1, a_2, \ldots,s_{T-1}, a_T, r_T, s_T}
\end{equation}

\subsection{Environment}

We adopt the discrete environment setting as the original R2R\cite{anderson2018vision} dataset proposed. In the discrete environment, the whole scene is modeled as a connectivity graph $\mathcal{G}$ with each node $\mathcal{V}$ representing a fixed location (also called navigable viewpoint), and the agent is only allowed to move between nodes. 
\begin{equation}
    \mathcal{G} = \{ \mathcal{V}, \mathcal{E} \}
\end{equation}
Where $\mathcal{E}$ represents edges between nodes $\mathcal{V}$; and each $e=(v_s, v_d) \in \mathcal{E}$ denotes a direct path between source viewpoint $v_s$ and destination viewpoint $v_d$. 

\subsection{Panoramic Action Space}
\label{sec: action space}

Following \cite{fried2018speaker} we use the panoramic action space where each viewpoint is represented as a panorama comprised of 3x12 views. Each view corresponds to a specific heading and elevation angle with FOV of $60$ degrees and resolution of 640x480. The heading angles range from $0$ to $360$ degrees, in $30$ degree intervals, and the elevation angles are $0, 30$ or $-30$ degrees. Among these 36 views, there are several views facing another viewpoint. Those views are called the candidate views. Once the agent turns to a candidate view, it can make a forward action and move to the corresponding viewpoint. The action decision process in the panoramic action space is to select candidate views consecutively for each time step until the agent chooses to stop. Generally, we use pre-extracted view features of a ResNet-152 model pre-trained on the Place365 dataset, so each view is represented as a single channel tensor $f_I$ of dimension $D_I = 2048$. 

Because the 36 views are represented as a set of vectors, it is important to retain the positional information for each view. Following \cite{anderson2018vision}, we use a quaternion of $(\mathrm{sin}(\psi_i), \mathrm{cos}(\psi_i), \mathrm{sin}(\omega_i), \mathrm{cos}(\omega_i))$ repeated by 32 times as the angle feature $f^i_A$ of the $i$ th candidate view, where $\psi_i, \omega_i$ denotes its relative heading and elevation to agent's current orientation. Its dimension $D_A = 128$. The angle feature of each view will be concatenated with the image feature to form the $i$ th view feature $\mathbf{V}^i = [f_I^i; f_A^i]$ of dimension $D_V = 2176$.



\subsection{States}

There are two states in the VLN task: the state of navigation and the state of the agent. 

The state of navigation is composed of the history of visual perceptions and the embedding of language instructions. This state is commonly maintained by a stateful module like an LSTM cell which will be updated during navigation. When we refer to states of the VLN model, we are generally talking about this one.

The state of the agent includes its pose, which is the embedding of its current heading and elevation $\{ \psi_t, \omega_t \}$. This orientation information is key to determine which view the agent is facing, and to better map commands like "turn left" in instruction to actions. 


\section{Method}

\subsection{Revisiting  Recurrent VLN-Bert}

We first revisit our transformer-based backbone: Recurrent VLN-Bert\cite{Hong2021vlnbert} before presenting the architecture of our model. 

In Recurrent VLN-Bert, the language instructions $\mathbf{L}_{in}$ are first tokenized as $\mathbf{L}_{tok}$ then fed into a BERT model to produce the instruction embeddings $\mathbf{L}$:
\begin{equation}
    \mathbf{L_{tok}} = [CLS]\mathbf{tokenizer}(\mathbf{L_{in}})[SEP]
\end{equation}
\begin{equation}
    \mathbf{L} = \mathrm{BERT}(\mathbf{L}_{tok})
\end{equation}

The embedding of the [CLS] token $\mathbf{L}_{cls}$ is treated as the initial navigation state. It will be updated during navigation, while the embeddings of the rest of the language tokens $\mathbf{L}_{/cls}$ remain fixed once being calculated.

In the action selection stage, a transformer decoder is used to perform multi-layer cross-attention between different modalities, and self-attention within each modality. In the multi-layer cross-attention module, the view features $\mathbf{V}_t$ concatenated with the state embedding $\mathbf{L}_{cls}$ are treated as queries. And the embeddings of the instruction without the [CLS] token $\mathbf{L}_{/cls}$ are served as the context (keys and values). 
\begin{equation}
    \mathbf{L}^{X}_{cls}, \mathbf{V}^{X}_t  = \mathrm{Attention}_{cross}([\mathbf{L}_{cls};\mathbf{V}_t], \mathbf{L}_{/cls})
\end{equation}
The superscript $X$ denotes the cross-attention. $\mathbf{V}_t = \{[f_I^i; f_A^i]\}_{i=0}^K$ where the subscript $t$ denotes the current time step, and the superscript $K$ denotes the number of candidate views in the current viewpoint (Note that there is always an additional special candidate view representing "stop"). 
The outputs of the cross-attention module are divided into two parts as $[\mathbf{L}^{X}_{cls}; \mathbf{V}^{X}_t]$. $\mathbf{L}^{X}_{cls}$ denotes the embedding of the [CLS] token and and $\mathbf{V}^X_t$ denotes the embeddings of visual inputs.

Then a multi-layer self-attention module will integrate information within candidate views. Also, visual information will be merged into the state embedding via $\mathbf{L}^{X}_{cls}$:
\begin{equation}
    \mathbf{L}_{cls}^{S}, \mathbf{V}^S_t = \mathrm{Attention}_{self}([\mathbf{L}_{cls}^X; \mathbf{V}^X_t])
\end{equation}
where the superscript $S$ denotes the self-attention. The outputs of the self-attention module are also divided into two parts like the cross-attention module. The attention weights from $\mathbf{L}_{cls}$ to the candidate views of the last self-attention layer are treated as the action selection score. The agent will choose the candidate view with the highest score to turn to and make a forward move or stop. 

Finally, a dot product is performed on the attended language embedding $\mathbf{L}_{/cls}^{X}$ and the attended visual embedding $\mathbf{V}^S_t$ to produce a joint representation $\mathbf{F}_{V, L}$ of two modalities. Then the state embedding is updated by a linear projection:
\begin{equation}
    s_t = \mathbf{W}_s [\mathbf{L}_{cls}^{S}; \mathbf{F}_{V, L}] + \mathbf{b}_s
\end{equation}
where $\mathbf{W}_s$ and $\mathbf{b}_s$ are weight and bias terms. Layer normalization is applied on both $\mathbf{F}_{V, L}$ and $s_t$, and we ignore these symbols for a clear notation. The state $s_t$ will replace the $\mathbf{L}_{cls}$ in language embeddings in the next time step to forward the state information.

\subsection{Slot Attention}

The Recurrent VLN-Bert\cite{Hong2021vlnbert} uses only candidate views as visual inputs to reduce the computational cost. However, by ditching the non-candidate views, the agent's visual perception is restricted to isolated candidate views. This harms the agent's ability to visually perceive the entire viewpoint because some objects that are referred to in the instructions may be partially or completely absent or obscured in candidate views. Consequently, the information of those objects can not be restored without more visual contexts from non-candidate views nearby.

To address this issue, we proposed a modified version of slot attention that enhances the features of candidate views with the features of panoramic views. The slot attention is first proposed in \cite{FrancescoLocatello2020ObjectCentricLW} which is a iterative attention-based module aiming to produce object-centric representations of the input image. First, $K$ vectors called slots are initialize randomly as queries for the slot attention module, the number of slots differs from different tasks. In each iteration, these slots will attend to $N$ feature maps of the input image. The feature maps are outputs of CNN-like backbones. The slots will then update themselves using the attention output with a skip connection. Then the updated slots will be used as new queries in the next iteration. Finally, the slots will become a task-dependent representation that captures different salient parts of the input image.

In the VLN task, we use a panoramic action space as described in section \ref{sec: action space}. Hence, it is natural to consider the panorama of each viewpoint as the "input image". As we already use pre-extracted features of views, we stack the features of all 36 panoramic views of a viewpoint as the feature maps of the panorama. Also, we want to maintain the positional information of each view, so we further concatenate the angle feature of each view to its image feature. In this way, the feature maps are represented as stacked view features of all panoramic views $\mathbf{V}_p = \{[f_I^i; f_A^i]\}_{i=0}^{35}$ of shape (36, 2176). And we treat the feature maps as input keys $I_{key}$ in our slot attention module.

Furthermore, as candidate views represents the navigable directions, they should be the most salient part of the viewpoint for a navigation task. With this prior knowledge, we initialize the slots with view features of the candidate views $\mathbf{V}_t$. These slots are used as input queries $I_{query}$ in our slot attention module.  

One more problem is that we want to update visual information only for the slots, rather than positional information. So we take out the angle feature part from view features and use only the image features $\{f_I^i\}_{i=0}^{35}$ as input values $I_{value}$ in our slot attention module. The overall module is described in Algorithm \ref{alg: slot-attn} in pseudo-code. 

\begin{algorithm}[htbp]
\caption{Our proposed slot attention module. At each viewpoint, the slots are initialized using features of candidate views as queries. The keys and values are initialized using feature of 36  panoramic views. Angle features are included in query and keys, but not in values. We set the number of iterations $T$ to 3, at each iteration the angle feature part of queries and keys are kept fixed.}
\label{alg: slot-attn}
\begin{algorithmic}[1]
\State \textbf{Input:} $\mathsf{I}_{query} \in \mathbb{R}^{K\times D_V}, \mathsf{I}_{key}\in \mathbb{R}^{N\times D_V}, \mathsf{I}_{value}\in \mathbb{R}^{N\times D_I}$
\State \textbf{Layer params:}
    \Statex \quad $\mathsf{k, q, v}$: linear projections for attention;
    \Statex \quad $\mathrm{GRU; MLP}$; 
    \Statex \quad $\mathrm{LayerNorm}$ (x3): only applies to image features;
    \Statex \quad $\mathrm{DropFeat}$: dropout that only applies to image features
    \Statex \quad $\mathrm{Mask}$: Image attentioin mask matrix
\State $\mathsf{slots} = \mathrm{DropFeat}(\mathsf{I}_{query})$
\State $\mathsf{K} = \mathsf{k}(\mathrm{DropFeat}(\mathrm{LayerNorm}(\mathsf{I}_{key})))^T$
\State $\mathsf{V} = \mathsf{v}(\mathrm{DropFeat}(\mathrm{LayerNorm}(\mathsf{I}_{value})))^T$
\For{t in 0\dots T} 
    \State $\mathsf{slots\_prev} =\mathsf{slots}[\dots, :D_I]$
    \State $\mathsf{slots} = \mathrm{LayerNorm}(\mathsf{slots})$
    \State $\mathsf{attn} = \mathrm{Softmax}(\mathrm{Mask}(\frac{1}{\sqrt{D_V}}\mathsf{K}\cdot \mathsf{q}(\mathsf{slots})^T), axis=\mathsf{slots})$
    \State $\mathsf{updates} = \mathsf{attn}\cdot \mathsf{V}$
    \State $\mathsf{gru\_state} = \mathrm{GRU}(state=\mathsf{slots\_prev}, inputs=\mathsf{updates})$
    \State $\mathsf{slots}[\dots, :D_I] += \mathrm{MLP(LayerNorm(}(\mathsf{updates}))$
\EndFor
\end{algorithmic}
\end{algorithm}

The slots will be refined at each iteration $t = 1...T$ via a softmax-attention operation. The softmax operation over slots ensures that each slot competes for all panoramic views. We update the image feature part only and keep the angle feature part fixed to keep the position information intact during slot attention. To further prevent the angle feature being corrupted, all layer normalization and dropout operations are performed on image feature part only. We call this kind of dropout operation as $\mathrm{DropFeat}$. For example:
\begin{equation}
    \mathrm{LayerNorm}(\mathsf{I}_{key}) = \{[\mathrm{LayerNorm}(f_I^i);f_A^i]\}_{i=0}^{35}
\end{equation}
\begin{equation}
    \mathrm{DropFeat}(I_{query}) = \{[\mathrm{Dropout}(f_I^k);f_a^k]\}_{k=0}^{K-1}
\end{equation}
Finally, the outputs of the slot attention module will be used to update the features of the candidate views in a residual paradigm. Note that we update the image features only and keep the angle features fixed:
\begin{equation}
    \hat{f}_I^k = f_I^k + \mathsf{slots}_T^{k}[\dots,;D_I]
\end{equation}
Where the subscript $I$ denotes "image". The superscript $k$ represents the $k$ th candidate view. $\mathsf{slots}_T^{k}$ is the updated slots of the last iteration. The set of updated features of the candidate views $\hat{\mathbf{V}}_t = \{[\hat{f}_I^k; f_a^k]\}^{K}_{k=0}$ will be fed into the action decoder as visual inputs. We demonstrate that by introducing our proposed slot attention module, the model converges faster and achieves better performance than the baseline model.

\subsection{Local Attention Mask}
\label{sec:local attention}

One remaining problem is that for two views that are spatially distant, it is less informative or even disruptive to exchange information between them via attention mechanism. Intuitively, we want the attention weights to be higher if the slot (candidate view) is closer to the context view, and lower if the slot and the context view are distant. Thus, we adopt a simple local attention mask to restrict the attention span, making the slot attention more stable.

\begin{figure}[htbp]
  \centering
  \includegraphics[width=\linewidth]{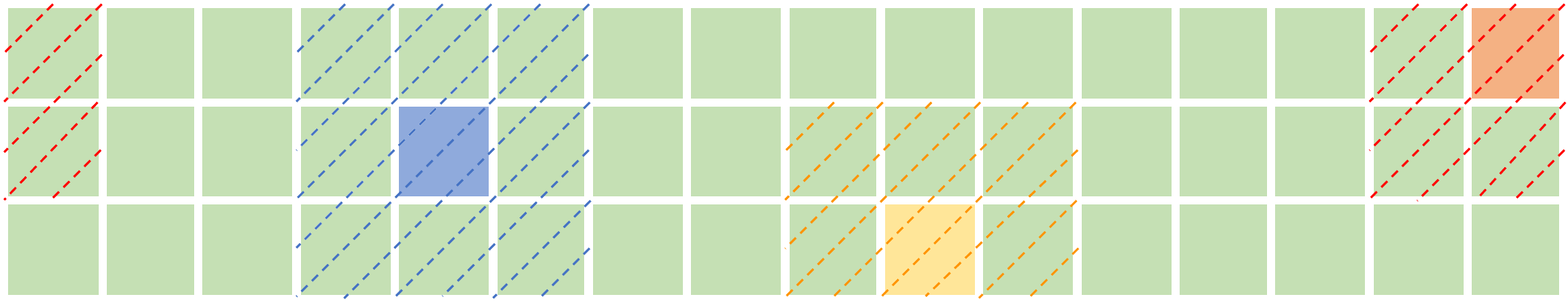}
  \caption{An example of a 3x3 local attention mask on a 3x12 grid of panoramic views. The grid is circular in the horizontal direction, three candidate views are filled with different colors than other background views. Only those views covered by the same color dashed line can be attended by each candidate view.}
  \Description{A 3x12 grid of panoramic views with a 3x3 local attention mask we propose.}
  \label{fig:mask}
\end{figure}

In practice, the 36 views are treated as a 3x12 grid. We apply a local mask on the regions centered on each candidate view to limit the attention span. For example, by applying a 3x3 local mask, only the 3x3 region around the candidate view can be attended by the candidate view. For candidate views in the top or bottom row of the grid, only a 2x3 region can be attended. An example of 3x3 mask is shown in Fig.\ref{fig:mask}.

In our experiment, we notice that candidate views tend to query views parallel to or below themselves. This makes sense because in the real world, most informative indoor landmarks are in the middle or lower parts of the environment, while the upper parts are usually ceilings only. This indicates that our slot attention module is capable of spotting the right regions that help navigation.

Our ablation study shows that applying the local attention mask mechanism to the slot attention module effectively improves the generalization ability of the agent.


\section{Experiments}

\newcolumntype{h}{>{\columncolor[RGB]{204, 255, 255}}c}
\begin{table*}[!htbp]
    \centering
    \scalebox{1.0}{
    \begin{tabular}{lccchccchccch}
        \hline
         \toprule
         \multicolumn{1}{c}{} & \multicolumn{4}{c}{Val Seen} & \multicolumn{4}{c}{Val Unseen} & \multicolumn{4}{c}{Test Unseen} \\
         \cmidrule(lr){2-5} \cmidrule(lr){6-9} \cmidrule(lr){10-13} 
         Agent & TL$\downarrow$ & NE$\downarrow$ & SR$\uparrow$ & SPL$\uparrow$ & TL$\downarrow$ & NE$\downarrow$ & SR$\uparrow$ & SPL$\uparrow$ & TL$\downarrow$ & NE$\downarrow$ & SR$\uparrow$ & SPL$\uparrow$ \\
         \hline
         RANDOM~\cite{anderson2018vision} & 9.58 & 9.45 & 16 & - & 9.77 & 9.23 & 16 & - &  9.93 & 9.77 & 13 & 12 \\
         Seq-to-Seq~\cite{anderson2018vision}& 11.33 & 6.01 & 39 & - & \textbf{8.39} & 7.81 & 22 & - & \textbf{8.13} & 7.85 & 20 & 18 \\
         \hline
         Look Before You Leap~\cite{wang2018look} & - & 5.56 & 43 & - & - & 7.65 & 25 & - & 9.15 & 7.53 & 25 & 23 \\
         Speaker-Follower~\cite{fried2018speaker} & - & 3.36 & 66 & - & - & 6.62 & 35 & - & 14.82 & 6.62 & 35 & 28 \\
         Self-Monitoring~\cite{ma2019self} & - & - & - & - & - & - & - & - & 18.04 & 5.67 & 48 & 35 \\
         Reinforced Cross-Modal~\cite{wang2019reinforced} & \textbf{10.65} & 3.53 & 67 & - & 11.46 & 6.09 & 43 & - & 11.97 & 6.12 & 43 & 38 \\
         FAST~\cite{ke2019tactical} & - & - & - & - & 21.17 & 4.97 & 56 & 43 & 22.08 & 5.14 & 54 & 41 \\
         ALTR~\cite{huang2019transferable} & 13.2 & 4.68 & 56 & 53 & 9.8 & 5.61 & 46 & 43 & 10.3 & 5.49 & 48 & 45 \\
         EnvDrop~\cite{tan2019learning} & 11.00 & 3.99 & 62 & 59 & 10.70 & 5.22 & 62 & 48 & 11.66 & 5.23 & 51 & 47 \\
         The Regretful Agent~\cite{ma2019regretful} & - & 3.23 & 69 & 63 & - & 5.32 & 50 & 41 & 13.69 & 5.69 & 48 & 48 \\
         AuxRN~\cite{zhu2020vision} & - & 3.33 & 70 & 67 & - & 5.28 & 55 & 50 & - & 5.15 & 55 & 51 \\
         PREVALENT~\cite{hao2020towards} & 10.32 & 3.67 & 69 & 65 & 10.19 & 4.71 & 58 & 53 & 10.51 & 5.30 & 54 & 51 \\
         PRESS~\cite{li2019robust} & 10.35 & 3.09  & 71 & 67 & 10.06 & 4.31 & 59 & 55 & 10.52 & 4.53 & 57 & 53 \\
         VLN $\circlearrowright$ BERT\cite{Hong2021vlnbert} & 11.13 & 2.90 & 72 & 68 & 12.01 & 3.93 & \textbf{63} & 57 & 12.35 & 4.09 & 63 & 57 \\
         \hline
         \textbf{LSA (Ours)}  & 11.25 & \textbf{2.51} & \textbf{76} & \textbf{72} & 11.87 & \textbf{3.92} & 62 & \textbf{57} & 12.80 & \textbf{4.07} & \textbf{64} & \textbf{59} \\
         \bottomrule
    \end{tabular}
    }
    \caption{Results comparing \textbf{OUR MODEL} with the previous state-of-the-art on the R2R val and test splits. The primary metric is highlighted in the light cyan color. }
    \label{tab:leaderboard}
\end{table*}

\subsection{Setup}

\subsubsection{Datasets}

We evaluate our proposed model on the Room-to-Room(R2R)\cite{anderson2018vision} dataset. The R2R dataset contains 10,567 panoramic views of 90 real-world in-door environments and 7,189 trajectories, each one is attached with three instructions written in natural language. Because the viewpoints of each environment are defined as nodes in a connectivity graph, the agent can navigate between viewpoints in the discrete space, that is to say, the agent can rotate in place or move forward to a linked viewpoint without collision. The vision-and-language task on the R2R dataset is to test the agent's ability to navigate in unfamiliar environments following natural language instructions.

\subsubsection{Evaluation Metrics}

We apply the same evaluation metrics used in existing works: (1) Navigation Error(\textbf{NE}), the shortest path length between the agent's final position and the goal location. (2) Success Rate (\textbf{SR}), the percentage of the agent successfully stops within 3 meters of the goal location. (3) Oracle Success Rate (\textbf{OSR}), as long as the agent is ever within 3 meters of the goal location in its trajectory, it is considered successful. (4) Success rate weighted by Path Length (\textbf{SPL}), trades-off success rate against trajectory length, higher \textbf{SPL} means more efficient navigation and less exploration.

\subsubsection{Implementation Details}.
We use the image features provided in R2R\cite{anderson2018vision} which are encoded by a ResNet-152\cite{he2016deep} pre-trained on Place365\cite{zhou2017places}. For the slot attention module, we set the dropout rate to 0.7, and the local attention mask shape to 3x3. In addition, for the "stop" view that indicates a stop action for each viewpoint, we use a max-pooled feature over all 36 panoramic views of that viewpoint as its feature. All experiments are conducted on a single NVIDIA 2080Ti GPU. We use an AdamW optimizer\cite{loshchilov2017decoupled} and a cosine annealing scheduler with warm-up\cite{loshchilov2016sgdr} to adjust the learning rate during training. The batch size is set to 8. We train the agent on both the original training data and the augmented data from PREVALENT\cite{hao2020towards} for 100K iterations.

\subsection{Main Results}
We compare our model with recently published SoTA models on R2R dataset. Results are demonstrated in Table.\ref{tab:leaderboard}. The most primary metric is the \textbf{SPL} (\textbf{S}uccess rate weighted by \textbf{P}ath \textbf{L}ength) and we highlight the corresponding column with a light cyan background. Our model are based on the Recurrent VLN-Bert\cite{Hong2021vlnbert} backbone.

Clearly our model achieves a performance boost on the validation seen data comparing with all previous works. We outperform our baseline Recurrent VLN-Bert\cite{Hong2021vlnbert} by a large margin of 4\% on both SR and SPL. This indicates that our proposed slot attention module can highly improve the agent's ability to memorize environments it has seen even we apply a large dropout rate of 0.7. Also, we achieve the lowest navigation error which indicates that our agent has a better comprehension on when and where to stop. 

Notice that we do not corrupt the generalization ability of the agent, for we achieve the same SPL and a slightly lower navigattion error on the validation unseen set. And we achieve these results on only 100K training iterations while the baseline Recurrent VLN-Bert\cite{Hong2021vlnbert} has been trained for 300K iterations. This shows that our model converges faster than the baseline model without performance loss.

Furthermore, on the test set where the environments and instructions are unseen during training, our model surpasses the Recurrent VLN-Bert\cite{Hong2021vlnbert} by 1\% on SR and 2\% on SPL. Again, the navigation error is the lowest among all models. These results show that our model has better generalization ability on the test set which has a significant different data distribution than the training and the validation set.



\subsection{Ablation Study}

We conducted several ablation studies on different components of our model. All models are trained for 100K iteration following the same setting as our full model. We retrain the Recurrent VLN-Bert\cite{Hong2021vlnbert} model as the baseline.

\subsubsection{Local mask.} We report the results of applying different shapes of the local attention mask on validation sets in the Table\ref{tab:ablation mask}. We find that the 3x3 local mask produces the most promising result. Comparing line 2 to line 4, we witness that a larger horizontal span produces better results on seen environments but somehow downgrade the agent's generalization ability to unseen environments. By expanding the vertical span from 1 to 3 as shown in lines 5 to 7, the generalization ability rises while the capability of memorizing seen environments declines. Two main findings can be witness from these results. One is that too large horizontal span truly introduces noise from distant views and harms the navigation, comparing line 3 to line 4 and line 6 to line 7. The other is that information of neighbor views in vertical directions actually matters for navigation, comparing line 2 to 5.
\begin{table}
    \caption{Ablation study on shape of the local attention mask.}
    \label{tab:ablation mask}
    \begin{tabular}{cccccccc}
         \toprule
         \multicolumn{1}{c}{} & \multicolumn{1}{c}{} & \multicolumn{2}{c}{Val Unseen} & \multicolumn{2}{c}{Val Unseen} \\
         \cmidrule(lr){3-4} \cmidrule(lr){5-6} 
         model & Mask Shape  & SR & SPL & SR & SPL \\
         \hline
         1 & none & 71.69 & 66.60 & 61.94 & 55.99 \\
         2 & 1x3 & 74.14 & 69.89 & 61.30 & 56.20 \\
         3 & 1x5 & 75.32 & 70.77 & 62.20 & 56.36 \\
         4 & 1x7 & 75.22 & 70.87 & 61.69 & 55.92 \\
         5 & 3x3 & \textbf{76.30} & \textbf{71.89} & \textbf{62.49} & \textbf{57.13} \\
         6 & 3x5 & 70.13 & 65.20 & 62.37 & 56.64 \\
         7 & 3x7 & 69.05 & 64.45 & 62.03 & 56.26 \\
         \bottomrule
    \end{tabular}
\end{table}
 

\subsubsection{Iterations of slot attention.}
We examine the effect of the iteration number of slot attention by setting it from 1 to 4. The results are shown in Table\ref{tab:ablation iter}. The first line with iteration 0 is the baseline model without slot attention module equipped. The result shows that the SPL on unseen set grows with the increase of the iteration number and reaches its peak on the iteration number of 3. Then it drops when the iteration number grows to 4. These results indicate that the multi-iteration slot attention does improve the model's performance, but too many iterations can make the model too complex to converge.
\begin{table}
    \caption{Ablation study on the number of iterations of the slot attention module.}
    \label{tab:ablation iter}
    \begin{tabular}{cccccccc}
         \toprule
         \multicolumn{1}{c}{} & \multicolumn{1}{c}{} & \multicolumn{2}{c}{Val Unseen} & \multicolumn{2}{c}{Val Unseen} \\
         \cmidrule(lr){3-4} \cmidrule(lr){5-6} 
         model & iters  & SR & SPL & SR & SPL \\
         \hline
         0 & 0 & 69.54 & 65.25 & 61.60 & 56.02 \\
         1 & 1 & 74.44 & 70.14 & 61.56 & 55.99 \\
         2 & 2 & 72.18 & 68.07 & 61.86 & 56.27 \\
         3 & 3 &\textbf{76.30} & \textbf{71.89} & \textbf{62.49} & \textbf{57.13} \\
         4 & 4 & 73.85 & 69.64 & 61.22 & 55.46 \\
         \bottomrule
    \end{tabular}
\end{table}




\subsection{Visualization}

\begin{figure*}
  \centering
  \includegraphics[width=0.88\linewidth]{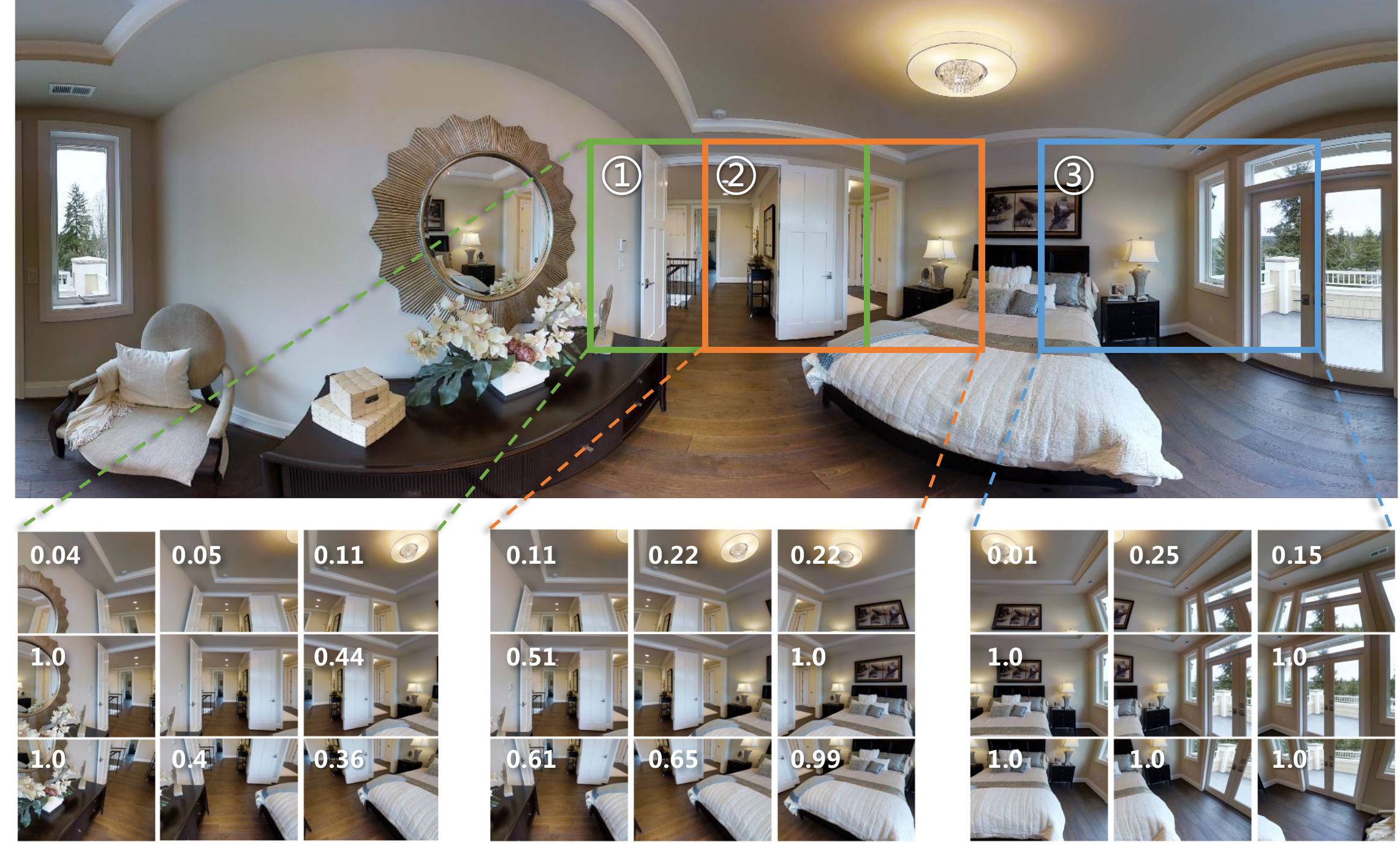}
  \caption{Instruction is \textit{"Go forward past the bed and the stairs. "}. On the top line, colored boxes with indexes indicate candidate views. Each candidate view connects to its local 3x3 regions on the bottom line with dashed lines. Values on the top left of each neighbor view are the attention weights from the candidate view to the neighbor view. In this case, baseline Model chooses \textcircled{\small{2}}, our model chooses \textcircled{\small{1}}.}
  \Description{An example showing our proposed \textbf{LSA} module with local masks can integrate information from neighbor views.}
  \label{fig:exp1}
\end{figure*}

As illustrated in Fig.\ref{fig:teaser}, candidate views can merge information from their neighbor views within a local region via the slot attention module. The higher opacity indicates higher attention weight from each candidate view to its surrounding context views. Here are more examples:

As shown in Fig.\ref{fig:exp1}, the top line is the panoramic view of one viewpoint consisting of 3x12 views. Candidate views are marked with colored square boxes and a circled number on the top left as their indexes. The bottom line is the 3x3 neighbor region of each candidate view, values on the top left are the attention weights of each candidate view to each of its neighbor views in slot attention. In this case, the sub-instruction should be \textbf{"Go forward past the bed and the stairs. "} in current time step. So the agent should focus on "bed" and "stairs". As illustrated, there are stairs in the center of the candidate view \textcircled{\small{1}}, but no bed can be seen. And for candidate view \textcircled{\small{2}}, the situation is the opposite. For the baseline model that only spots isolated candidate views, it wrongly selects candidate view \textcircled{\small{2}} to be its next move. However, our model can integrate information from neighbor views so it captures the "bed" to the right of the candidate view \textcircled{\small{1}} and makes the correct move.

\begin{figure*}
  \centering
  \includegraphics[width=0.88\linewidth]{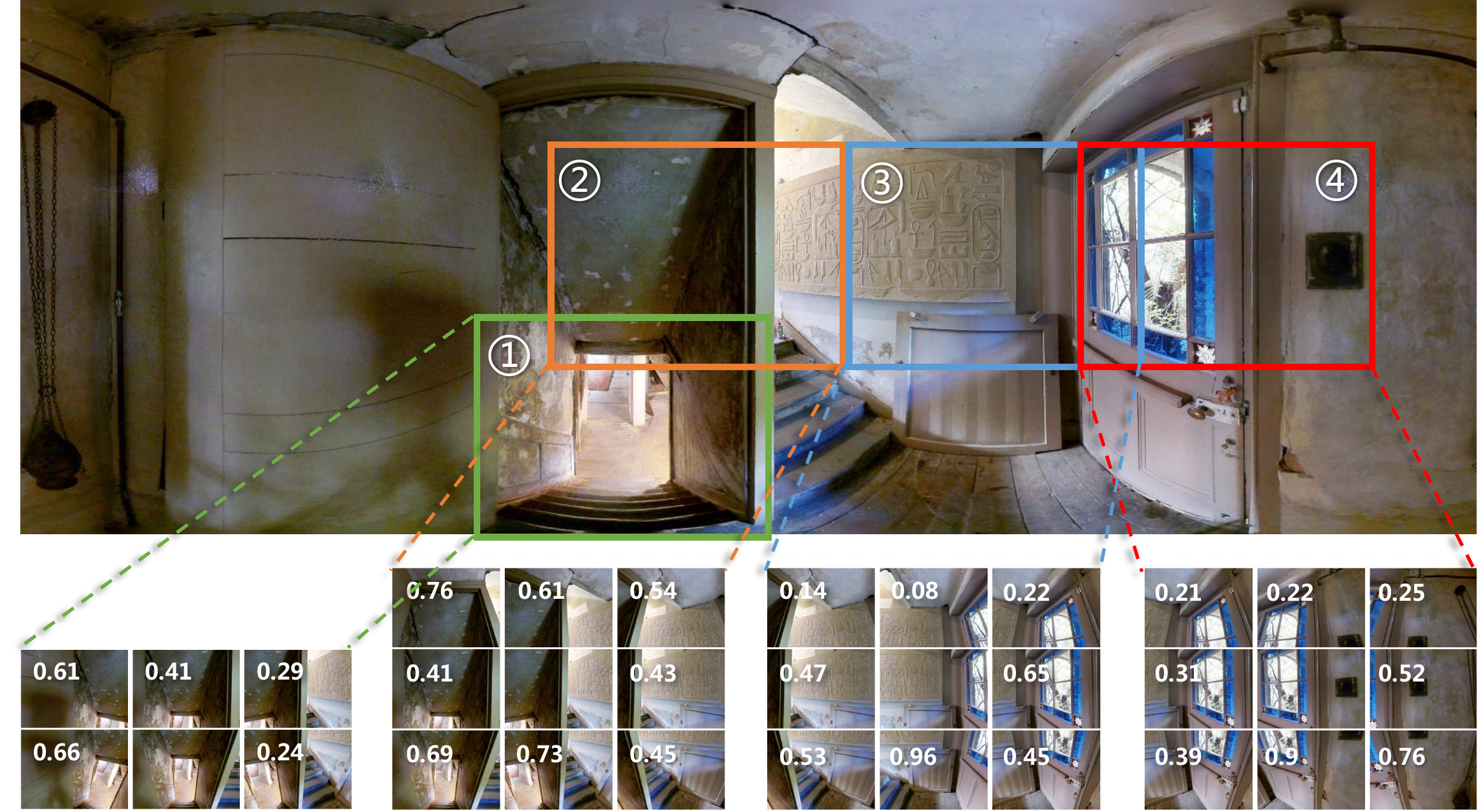}
  \caption{Instruction is \textit{"Turn into the first door on the right."}. In this case, baseline Model chooses \textcircled{\small{2}}, our model chooses \textcircled{\small{4}}. Notice that candidate view \textcircled{\small{1}} is at the bottom line with an elevation of -30 degrees. Thus, only a 2x3 mask is applied for this view which means the views at the top line with an elevation of 30 degrees will not be attended by candidate view \textcircled{\small{1}}.}
  \Description{Another example showing our proposed \textbf{LSA} module helps the agent to restore information of partially obscured objects.}
  \label{fig:exp2}
\end{figure*}

Another example is in Fig.\ref{fig:exp2} where the sub-instruction in the current time step should be \textbf{"Turn into the first door on the right."}. From the panoramic perspective, we can identify that the candidate view \textcircled{\small{4}} is part of a door to which the instruction refers. However, it can be hard to tell if it is a window depending on the single candidate view \textcircled{\small{4}} only. So baseline model wrongly chooses candidate view \textcircled{\small{2}} while our model correctly chooses candidate view \textcircled{\small{4}}. Notice that the view under candidate view \textcircled{\small{4}} represents a key perspective for identifying the complete "door", so it gains a large attention weight of 0.9.

\section{Conclusion}
In this paper, we present a slot-attention based module to adaptively incorporate information from panoramic views to candidate views. By applying this module with a local attention mask, our agent is equipped with a larger visual perceptive field that helps the agent to better comprehend each navigable region. Extensive experiments demonstrate that our model effectively boosts the performance on seen data while preserving a great generalization ability. And it is noteworthy that our model achieves the state-of-the-art in one-third training iterations compared with the baseline model.

Considering the object information can be useful to navigation, we believe that explicitly introducing object features into inputs of our proposed slot attention module can further improve the performance, since object features may better match the language feature than the raw image feature. We leave this as future works to explore.

~\\

\noindent \textbf{Acknowledgment}.
This work was supported by the Science and Technology Major Project of Commission of Science and Technology of Shanghai(No.21XD1402500)







\bibliographystyle{ACM-Reference-Format}
\balance
\bibliography{icmr}

\end{document}